\documentclass[11pt]{article}
\usepackage[preprint]{acl}
\usepackage{times}
\usepackage{latexsym}
\usepackage[T1]{fontenc}
\usepackage[utf8]{inputenc}
\usepackage{microtype}
\usepackage{inconsolata}

\usepackage{amsmath,amssymb}
\usepackage{booktabs}
\usepackage{graphicx}
\usepackage{xcolor}
\usepackage{enumitem}
\usepackage{tabularx}
\usepackage{multirow}
\usepackage{algorithm}
\usepackage{algpseudocode}
\usepackage{tikz}
\usetikzlibrary{arrows.meta, positioning}

\setlist{nosep,leftmargin=*}

\newcommand{\system}{\textsc{Cargo}}
\newcommand{\bench}{\textsc{Cargo-Bench}}
\newcommand{\rid}{RID}

\newcommand{\stSup}{\textsc{sup}}
\newcommand{\stCon}{\textsc{con}}
\newcommand{\stUnv}{\textsc{unv}}

\title{\system: Context-Aware Retrieval-Gated Evaluation\\of Agentic AI in Production}

\author{Mukul Chhabra \and Shail Patel \and Luigi Medrano \\
  Dell Technologies}

\begin{document}
\maketitle

\begin{abstract}
Reference-based LLM-as-a-judge evaluation assumes that a reference answer is the target for the response under evaluation. In deployed agentic systems that operate over dynamic entities---support cases, assets, accounts---this assumption fails: the closest available reference typically instantiates the \emph{correct procedure} on a \emph{different entity}, so a judge that compares literally penalizes legitimately different identifiers, dates, and statuses as errors or hallucinations. We name this failure mode \emph{reference--instance divergence} (\rid) and show that it is structural rather than incidental. We propose \system, a framework that (i) treats retrieved references as procedural exemplars and grounds factual judgments in the live instance's observed context, (ii) assigns each claim a three-way status---supported, contradicted, or unverifiable---penalizing only contradictions with observed facts, and (iii) gates evaluation by retrieval confidence, casting continuous production evaluation as selective prediction with an explicit risk--coverage--cost trade-off. To measure the effect without relying solely on costly expert labels, we introduce \bench, a perturbation-based diagnostic suite whose ground truth holds by construction. \bench\ separates \emph{leniency} from \emph{discrimination}: a valid judge must stop penalizing entity transplants while still detecting injected contradictions and procedural corruptions. On \bench\ (246 items, two judge models, 7{,}872 judgments), the standard reference-based judge penalizes 100\% of correct entity-transplanted answers as incorrect and hallucinated---it is uninformative (DI $\approx 0$)---and supplying the live facts without reframing changes nothing. \system\ eliminates these false penalties (0/50 on transplants) while retaining near-complete contradiction recall (50/50 and 49/50), raising DI to .58 [.48, .68]; a rubric-swap control attributes most of the effect to context-grounded dimension definitions rather than prompt framing. \system\ also exposes a limitation of its own design: the same leniency that protects entity values suppresses detection of procedural corruptions (20\% recall), a trade-off DI makes visible; an explicit post-hoc fix targeting exactly this failure does not close the gap ($\Delta$DI $=-.007$ [$-.038,.023$]), and an LLM-as-annotator study with written guidelines and adjudication shows the same blind spot (3/10 procedural corruptions recovered). We release a preregistered protocol for extending the evaluation to expert agreement, risk--coverage, and cost on production traffic.
\end{abstract}

\section{Introduction}
\label{sec:intro}

LLM-as-a-judge has become the default instrument for evaluating open-ended model outputs at scale \citep{zheng2023judging,liu2023geval,gu2024survey}. In the reference-based variant, a judge is shown a candidate response together with a gold reference and asked to score the candidate's correctness, completeness, or faithfulness relative to that reference. This paradigm rests on an assumption so familiar that it is rarely stated: \emph{the reference is the target}. For static tasks such as summarization or knowledge QA the assumption is reasonable.

It fails for a large and growing class of deployed systems. Agentic assistants in technical support, customer service, IT operations, and finance answer questions \emph{about specific entities}---a particular support case, service tag, order, or account---by retrieving facts from backend systems and applying a domain procedure to them. Two users may ask the same question about different entities. The correct answers share a procedure and a structure but differ in almost every surface value. A curated golden set can realistically cover the space of \emph{questions and procedures}; it cannot cover the space of \emph{entities}. Consequently, at production time the nearest available reference is almost always an answer to the same question about a different entity.

We call this condition \emph{reference--instance divergence} (\rid). Under \rid, a literal reference-based judge systematically confuses two orthogonal properties: whether the response follows the correct procedure, and whether its entity-specific values agree with the reference's. The second comparison is meaningless---the values \emph{should} differ---yet it dominates a naive judge's verdict, depressing correctness scores and inflating hallucination flags. Section~\ref{sec:problem} formalizes this and shows why the effect is structural rather than a matter of prompt wording.

We further observe that in production, evaluation itself is a decision under uncertainty. Not every live interaction has an appropriate reference in the golden set, and a judge applied with an irrelevant reference produces a confidently wrong score. Evaluating every interaction is also expensive. Continuous production evaluation is therefore a \emph{selective prediction} problem \citep{elyaniv2010foundations,geifman2017selective}: the evaluator should abstain when it lacks a valid reference, and the appropriate objects of study are the risk--coverage curve and the cost curve, not a single accuracy number.

We present \system\ (\textbf{C}ontext-\textbf{A}ware \textbf{R}etrieval-\textbf{G}ated evaluati\textbf{O}n), which addresses both problems, and \bench, a diagnostic suite that makes the improvement measurable with ground truth that holds by construction. Our contributions are:

\begin{enumerate}
\item \textbf{Problem.} We identify and formalize reference--instance divergence, decomposing a response into a transferable procedure and instance-specific parameters, and show that literal reference comparison is not identifiable with respect to procedural correctness under \rid\ (\S\ref{sec:problem}).
\item \textbf{Method.} \system\ (\S\ref{sec:method}) reinterprets the retrieved reference as a \emph{procedural exemplar}, grounds factual judgment in observed live context with three-way claim status (\stSup/\stCon/\stUnv), and gates evaluation on retrieval confidence with calibrated and margin-aware variants. It evaluates the full agentic trace---routing, replanning, response type, latency---not only the final answer.
\item \textbf{Benchmark.} \bench\ (\S\ref{sec:bench}) uses five controlled perturbation families (entity transplant, contradiction injection, unverifiable augmentation, procedural corruption, retrieval distractors) to produce labeled evaluation instances by construction. It yields a \emph{discrimination index} that jointly penalizes false penalties and missed errors, so that a judge cannot score well merely by being lenient.
\item \textbf{Findings.} On \bench\ (246 items), literal reference judging is completely uninformative under \rid\ (DI $\approx 0$ on both judge models: every correct transplant is called wrong and hallucinated), and merely adding the live facts does not help. \system\ removes the false penalties without losing contradiction recall ($\Delta$DI $=+.58$ [.48, .68]). A rubric-swap control shows the effect is carried mostly by context-grounded dimension definitions, and per-family analysis reveals that the method's leniency over-generalizes from entity values to procedure (20\% recall on procedural corruptions); an explicit per-claim-type authority fix we test post-hoc does \emph{not} resolve this ($\Delta$DI $=-.007$ [$-.038,.023$]), so we characterize rather than patch it (\S\ref{sec:results}--\S\ref{sec:analysis}). We further specify a preregistered protocol for expert agreement, retrieval gating, and cost on production traffic (\S\ref{sec:setup}).
\end{enumerate}

\section{Reference--Instance Divergence}
\label{sec:problem}

\subsection{Setting}
A live interaction is a tuple $x=(q,a,c,o)$: a user question $q$, the system's answer $a$, an observed \emph{context} $c$ consisting of instance-specific facts available in the trace input (e.g., case number, subject, description, asset identifier), and a trace $o$ of observations (planner decision, agent calls, tool outputs, timings). A golden set $\mathcal{G}=\{(q^g_i,a^g_i,m^g_i)\}_{i=1}^N$ holds reference questions, reference answers, and metadata $m^g_i$ such as the expected intent or agent.

\subsection{Procedure--parameter decomposition}
We model an answer as the application of a procedure to instance parameters:
\begin{equation}
a = \pi(\theta) \oplus \epsilon,
\label{eq:decomp}
\end{equation}
where $\pi$ is a domain procedure (which checks to perform, which policy to apply, which fields to report, in which structure), $\theta$ is the vector of instance-specific values, and $\epsilon$ collects everything else (phrasing, formatting). A reference $a^g=\pi^g(\theta^g)\oplus\epsilon^g$ is written for its own instance $\theta^g$. The live answer should satisfy $\pi=\pi^g$ (correct procedure) and $\theta$ consistent with the live instance, \emph{not} with $\theta^g$.

\subsection{Non-identifiability of literal comparison}
A literal reference-based judge computes some divergence $D(a,a^g)$ and maps it to a score. Under the decomposition, $D$ mixes two terms,
\begin{equation}
D(a,a^g)\;\approx\;D_\pi(\pi,\pi^g)\;+\;D_\theta(\theta,\theta^g),
\end{equation}
and only the first is informative about quality. When \rid\ holds ($\theta\neq\theta^g$ by design), $D_\theta$ is large for \emph{every} correct answer. The judge therefore cannot distinguish a correct answer for a different instance from an incorrect one: correctness is not identifiable from $D$ alone. This is why prompt-level exhortations to ``focus on reasoning'' are insufficient in practice: the judge is given no signal with which to separate the two terms. The fix must supply that signal---the live instance's observed facts $c$---and must change what the reference is \emph{for}.

\subsection{Three-way claim status}
Let $\mathcal{K}(a)$ be the set of atomic factual claims in $a$ \citep{min2023factscore}. Relative to the observed context $c$, each claim $k$ has a status
\begin{equation}
\sigma(k\mid c)\in\{\stSup,\;\stCon,\;\stUnv\},
\end{equation}
supported if entailed by $c$, contradicted if inconsistent with a fact in $c$, and unverifiable otherwise. Because $c$ is a \emph{partial} snapshot of the request input---the agent legitimately retrieves many more fields from backend systems---the unverifiable class is large and expected. We define
\begin{align}
\mathrm{ContraRate}(a,c)&=\tfrac{|\{k:\sigma=\stCon\}|}{|\mathcal{K}(a)|},\\
\mathrm{UnvRate}(a,c)&=\tfrac{|\{k:\sigma=\stUnv\}|}{|\mathcal{K}(a)|}.
\end{align}
The central normative choice of \system\ is that hallucination is scored from $\mathrm{ContraRate}$ alone; $\mathrm{UnvRate}$ is reported as a coverage diagnostic rather than folded into a penalty. This choice trades recall on fabricated-but-unverifiable claims for precision on the claims that can actually be checked; \S\ref{sec:limitations} discusses the trade-off and \bench\ P3 measures it.

\section{The \system\ Framework}
\label{sec:method}

\begin{figure}[t]
\centering
\resizebox{0.85\columnwidth}{!}{%
\begin{tikzpicture}[
  box/.style={draw, rounded corners, align=center, font=\scriptsize, minimum height=8mm, minimum width=22mm, inner sep=1.3mm, fill=blue!4},
  small/.style={draw, rounded corners, align=center, font=\scriptsize, minimum height=8mm, minimum width=18mm, inner sep=1.3mm, fill=red!5, dashed},
  arr/.style={-{Latex[length=1.6mm]}, thick},
  lbl/.style={font=\tiny}
]
\node[box] (trace) at (0,6.2) {Trace};
\node[box] (embed) at (0,4.9) {Batch embedding};
\node[box] (gate)  at (0,3.6) {Retrieval gate};
\node[small] (abst) at (-2.6,2.0) {Abstain};
\node[box, minimum width=26mm, minimum height=11mm] (judge) at (2.6,2.0) {\system\ judge\\(exemplar +\\ctx-grounded)};
\node[box] (diag)  at (0,0.3) {Trace diagnostics};
\node[box] (rec)   at (0,-1.0) {Persisted record};

\draw[arr] (trace) -- (embed);
\draw[arr] (embed) -- (gate);
\draw[arr] (gate.south) -| node[lbl, pos=0.15, below, xshift=-2mm]{margin$<\tau$} (abst.north);
\draw[arr] (gate.south) -| node[lbl, pos=0.15, below, xshift=2mm]{margin$\geq\tau$} (judge.north);
\draw[arr] (abst.south) |- (diag.west);
\draw[arr] (judge.south) |- (diag.east);
\draw[arr] (diag) -- (rec);
\end{tikzpicture}%
}
\caption{Overview of \system. A live question is embedded and matched against the golden set; if the retrieval margin clears $\tau$, the retrieved reference is reinterpreted as a procedural exemplar and the answer is judged against it plus the live context; below $\tau$, \system\ abstains. Both paths log to trace diagnostics.}
\label{fig:overview}
\end{figure}
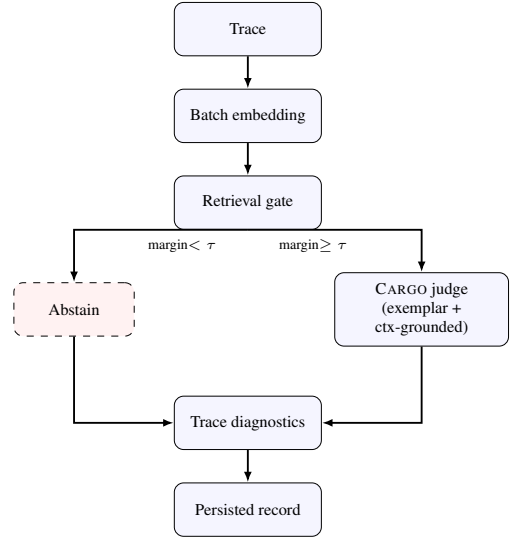

\system\ has four components (Figure~\ref{fig:overview}): a retrieval gate, a procedural-exemplar reinterpretation of the reference, a context-grounded structured judge, and trace-level diagnostics. Algorithm~\ref{alg:cargo} gives the per-interaction procedure.

\begin{algorithm}[t]
\caption{\system\ per-interaction evaluation}
\label{alg:cargo}
\begin{algorithmic}[1]
\Require interaction $x=(q,a,c,o)$; golden set $\mathcal{G}$ with cached embeddings $E^g$; gate $g$; judge $J$
\State $e\gets \mathrm{Embed}(q)$ \Comment{batched across a poll window}
\State $s_i\gets \cos(e,E^g_i)\;\forall i$;\; $i^\star\gets\arg\max_i s_i$
\State $\Delta\gets s_{(1)}-s_{(2)}$ \Comment{top-1/top-2 margin}
\If{$g(s_{i^\star},\Delta,m^g_{i^\star})=0$}
  \State \Return \textsc{abstain}$(s_{i^\star},\Delta)$
\EndIf
\State $r\gets \mathrm{ResponseType}(q,a)$; $p\gets \mathrm{PlannerEval}(o,m^g_{i^\star})$
\State $\ell\gets \mathrm{Latency}(o)$; $\rho\gets \mathrm{ReplannerEval}(o)$
\If{$r=\textsc{actual}$ $\wedge$ guards pass}
  \State $y\gets J(q,a,c,a^g_{i^\star})$ \Comment{exemplar + context}
  \If{$\neg\mathrm{Parses}(y)$} \Return \textsc{error} \EndIf
\EndIf
\State \Return record$(y,r,p,\rho,\ell,s_{i^\star},\Delta,i^\star,c)$
\end{algorithmic}
\end{algorithm}

\subsection{Retrieval gate}
\label{sec:gate}
Golden questions are embedded once and cached; live questions are embedded in batches per polling window to bound API calls and rate-limit pressure. With cosine similarity $s^\star$ to the nearest golden question and margin $\Delta=s_{(1)}-s_{(2)}$, we study three gates:
\begin{description}[style=nextline,leftmargin=0.4cm]
\item[Fixed.] $g=\mathbb{I}[s^\star\ge\tau]$. Operational default $\tau=0.80$, which must be recalibrated per embedding model.
\item[Margin-aware.] $g=\mathbb{I}[s^\star\ge\tau\wedge\Delta\ge\delta]$. Rejects ambiguous matches whose top two candidates are nearly tied, even when both exceed $\tau$.
\item[Calibrated.] Fit $\hat{P}(\text{reference appropriate}\mid s^\star,\Delta,\text{intent})$ by isotonic or logistic regression on a held-out calibration split labeled for reference appropriateness (\S\ref{sec:annotation}), and gate on $\hat{P}\ge\gamma$. This makes the operating point interpretable as a target reference-validity rate and admits per-intent thresholds.
\end{description}
The gate decides whether the \emph{evaluator} has a usable reference; it says nothing about whether the production answer is good. Abstentions are retained as coverage gaps for human review and golden-set expansion (\S\ref{sec:analysis}).

\subsection{Reference as procedural exemplar}
\label{sec:exemplar}
For accepted interactions the judge receives $q$, $a$, the observed context $c$, and the retrieved reference $a^g_{i^\star}$ with an explicit reinterpretation: the reference was written for a \emph{different} instance; it specifies the correct procedure, policy, required steps, and structure; its entity-specific values are not targets. The judge is asked to infer the reference's implicit instance from its text and to evaluate whether the live answer applies the same procedure \emph{correctly adapted} to the live instance. Appendix~\ref{app:prompts} gives the full template.

\subsection{Context-grounded structured judge}
\label{sec:judge}
The judge scores four dimensions on $[0,1]$, each defined against the live context rather than the reference: \textbf{correctness} (correct procedure; no claim contradicts $c$), \textbf{completeness} (covers the steps the exemplar's procedure requires, adapted to $c$), \textbf{helpfulness} (resolves $q$ clearly), and \textbf{hallucination fidelity} (no claim contradicts $c$; unverifiable claims are neutral). Output is schema-constrained JSON with a per-dimension rationale; unparseable or empty outputs are rejected rather than persisted, so aggregates are not biased by silent failures. Dimension definitions are configuration, not code, and are held fixed across all conditions in our experiments.

\subsection{Trace-level diagnostics}
Final-answer quality alone cannot localize a failure in an agentic workflow. \system\ additionally records response type (actual answer, clarification request, guardrail), planner verdict against the expected agent in $m^g_{i^\star}$, replanner behavior, per-observation and end-to-end latency, language and length guards, and all retrieval metadata ($s^\star$, $\Delta$, $i^\star$) and the context $c$ used. These fields support the attribution analysis in \S\ref{sec:analysis}.

\subsection{Continuous operation}
A polling or backfill cycle fetches traces newer than a watermark, deduplicates, extracts $(q,a,c,o)$ from heterogeneous trace envelopes, runs Algorithm~\ref{alg:cargo}, persists successful records, and advances the watermark, reporting seen/matched/evaluated/abstained/error counts. Evaluation cost per window of $M$ traces is
\begin{equation}
C(\tau)=M\,C_e+M\cdot\mathrm{Cov}(\tau)\cdot C_j,
\label{eq:cost}
\end{equation}
with embedding cost $C_e\ll C_j$ (judge cost), so cost is governed almost entirely by coverage.

\section{\bench: Ground Truth by Construction}
\label{sec:bench}

Expert annotation of production traces is necessary but slow, expensive, and---critically for our claim---cannot by itself distinguish a judge that is \emph{better} from one that is merely \emph{more lenient}. \bench\ addresses this with controlled perturbations of seed items whose labels hold by construction.

\paragraph{Seeds.} Each seed is a golden item $(q^g,a^g,\theta^g)$ with its entity parameters $\theta^g$ explicitly annotated (fields and values). Seeds span all intents in the deployment.

\paragraph{Perturbation families.} From each seed we generate instances with known target verdicts:
\begin{description}[leftmargin=2.0cm,style=multiline]
\item[P1 Transplant] (correct) Sample $\theta^\ell$, rewrite $a^\ell=\pi^g(\theta^\ell)$; $c\subset\theta^\ell$. Target: high correctness, no hallucination. Measures the \emph{false-penalty rate} under \rid.
\item[P2 Contradiction] (hallucinated) From a P1 instance, alter one value in $a^\ell$ to conflict with a fact \emph{present} in $c$. Target: flagged. Measures \emph{contradiction recall}.
\item[P3 Unverifiable] (neutral) From a P1 instance, add $k$ plausible fields absent from $c$. Target: no penalty. A labeled ``fabricated'' sub-split measures the recall cost of the \stUnv\ policy.
\item[P4 Corruption] (incorrect) From a P1 instance, delete a required step, swap the policy, or invert a conditional. Target: low correctness/completeness---the check leniency alone cannot pass.
\item[P5 Distractor] Paraphrases of $q^g$ with a different intent/procedure. Target: gate abstains or assigns low $\hat P$. Measures the gate's reference-validity discrimination.
\end{description}
Perturbations are generated by templated rewriting with an LLM and verified by a second model and by rule-based consistency checks (every occurrence of a swapped value is updated; injected contradictions target a field that is in $c$). An author spot-check of 40 instances (Appendix~\ref{app:bench}) additionally reads each rendered triple for semantic validity beyond the automated checks. Appendix~\ref{app:bench} lists templates.

\paragraph{Discrimination index.} Let $\mathrm{FP}$ be the fraction of P1$\cup$P3 instances a judge penalizes (correctness $<0.5$ or hallucination flagged), and $\mathrm{TP}$ the fraction of P2$\cup$P4 instances it penalizes. We report
\begin{equation}
\mathrm{DI}=\mathrm{TP}-\mathrm{FP}\in[-1,1],
\end{equation}
alongside the full confusion matrix. A lenient judge lowers FP but also TP; a valid judge must raise DI. \bench\ therefore converts our central claim into a single falsifiable quantity.

\section{Experimental Setup}
\label{sec:setup}

\subsection{Deployment and data}
\label{sec:data}
Experiments use a production multi-agent assistant for enterprise technical support in which a planner routes each request to one of several specialized agents (policy, case, work-order, task, asset). Traces are collected via an observability layer that records inputs, outputs, and per-observation timings. The deployment's curated golden set comprises 30 question--answer--intent items; 23 are entity-free policy answers (for which \rid\ does not arise) and the remainder concern specific cases, work orders, and assets. We draw a production sample of $M=400$ traces stratified by intent, retrieval similarity decile, and response type, oversampling low-margin retrievals and planner failures with a 150-item minimum quota for the entity-bearing (\rid-eligible) stratum, since H1's effect concentrates there. $M=400$ is powered to detect $\Delta\rho \geq 0.15$ between \textsc{Direct} and \textsc{Cargo} (Fisher $z$, dependent correlations, power $0.8$, $\alpha=0.05$) and matches typical budgets in published LLM-judge agreement studies (200--500 items). All data are de-identified before annotation (\S\ref{sec:ethics}).

\paragraph{\bench\ instantiation.}
\bench\ is generated from ten seeds in two batches: an initial six (three derived from the entity-bearing golden items---case status, work-order status, asset age/warranty---lightly templated so entity parameters are explicit, and three synthetic seeds covering case, warranty, and work-order procedures), plus four added seeds spanning the same real intents with different question styles (case priority/assignment, asset entitlement, work-order parts, and a task-status seed) to increase statistical power and intent diversity. With five draws per seed and family (generation seeds 2027 and 3141), this yields 296 instances that pass all consistency checks: 50 P1, 50 P2, 96 P3 (50 true-but-unobserved, 46 fabricated), 50 P4 (26 step deletions, 10 negations, 14 policy swaps), and 50 P5. Observed context contains 1.12 facts on average ($|c|\in\{1,2,3\}$); answers average 24 words. P2 contradictions target case numbers (31), service tags (15), and subjects (4).

\subsection{Expert annotation}
\label{sec:annotation}
Three domain experts annotate each sampled trace with: (a) whether the retrieved reference is an appropriate procedural exemplar (binary, used to fit the calibrated gate and for retrieval evaluation); (b) correctness, completeness, helpfulness, and hallucination on a 5-point ordinal scale defined against the live context, using the same definitions as the judge; (c) status $\sigma\in\{\stSup,\stCon,\stUnv\}$ for each disputed claim; (d) acceptability of the planner's routing. Two annotators label every item; a third adjudicates disagreements. We report Krippendorff's $\alpha$ per dimension and release guidelines (Appendix~\ref{app:guidelines}). A held-out 20\% of annotated items is used only for gate calibration and threshold selection; all headline numbers are computed on the remainder.

\subsection{Conditions}
\label{sec:conditions}
All judge conditions share the judge model, decoding parameters (temperature 0.4 unless varied), dimension definitions, and JSON schema; they differ only in the information and framing given to the judge.
\begin{description}[leftmargin=3.0cm,style=multiline]
\item[\textsc{NoRef}] Question, answer, rubric only.
\item[\textsc{Direct}] Standard reference-based judging: compare $a$ to $a^g$.
\item[\textsc{Direct+Ctx}] \textsc{Direct} with $c$ appended, but no exemplar reinterpretation or \stUnv\ rule. Isolates ``more information'' from ``different framing.''
\item[\textsc{Cargo}] Full method: exemplar framing, context grounding, \stUnv\ rule.
\item[\textsc{Cargo}$-$\stUnv] \textsc{Cargo} with binary claim status (absent $\Rightarrow$ false). Tests the \stUnv\ rule.
\item[\textsc{Cargo}$-$Ex.] \textsc{Cargo} without the ``different instance'' framing (\textsc{Cargo}$-$Exemplar in tables).
\item[\textsc{Cargo}$-$Ctx] \textsc{Cargo} with $c$ removed.
\item[\textsc{Oracle}] \textsc{Cargo} with the human-selected appropriate reference; upper-bounds gains attributable to retrieval.
\item[\textsc{RandRef}] \textsc{Cargo} with a random same-intent reference; lower-bounds the value of retrieval.
\end{description}
Retrieval baselines: BM25 \citep{robertson2009bm25}, dense top-1 with the deployed embedding model, a sentence-transformer alternative \citep{reimers2019sbert}, intent-filtered dense, and hybrid. Gating baselines: no gate; random sampling at matched coverage; fixed, margin-aware, and calibrated gates (\S\ref{sec:gate}). One additional condition, \textsc{Cargo}$+$\textsc{Auth} (an explicit per-claim-type authority rule), was added \emph{post-hoc} after observing \textsc{Cargo}'s P4 gap and is reported as exploratory, not preregistered (\S\ref{sec:analysis}).

\subsection{Judge models and bias controls}
We run every condition with the same two judge models used for \bench\ (\S\ref{sec:res-bench})---gpt-oss-120b (three seeds) and gpt-oss-20b (one seed), open-weight and distinct from the production assistant's generator, limiting self-preference \citep{panickssery2024selfrecognition}---so bench and production-sample results are directly comparable; a third judge family would further strengthen H5 but is left to future work rather than added post hoc. We report per-seed means and within-item dispersion. Because \system\ judges single responses, position bias \citep{wang2024fair} does not arise; we control for verbosity bias via agreement stratified by answer-length quartile and a length-controlled regression \citep{dubois2024lengthcontrolled}.

\subsection{Metrics}
\label{sec:metrics}
\textbf{Agreement:} Spearman $\rho$ and Pearson $r$ against adjudicated expert scores per dimension; quadratic-weighted Cohen's $\kappa$ after binning; paired bootstrap 95\% CIs (10k resamples); Holm correction across dimensions. \textbf{Hallucination:} with expert \stCon\ labels as positives, precision/recall/F1 and two targeted false-positive rates (cross-instance differences; \stUnv\ claims). \textbf{\bench:} per-family penalty rates, DI, and DI stratified by intent and by $|c|$. \textbf{Retrieval:} top-1 accuracy and MRR against expert reference-appropriateness labels; fraction of traces with no appropriate reference in $\mathcal{G}$. \textbf{Selective evaluation:} for gate $g$ and disagreement loss $\ell$,
\begin{align}
\mathrm{Cov}(g)&=\tfrac{1}{M}\textstyle\sum_j g(x_j),\\
\mathrm{Risk}(g)&=\tfrac{\sum_j g(x_j)\,\ell(\hat y_j,y_j)}{\sum_j g(x_j)},
\end{align}
risk--coverage curves, AURC \citep{geifman2017selective}, and stream-level failure recall (the fraction of \emph{all} expert-identified failures the gated system evaluates and flags, since abstaining on hard failures trivially lowers risk). \textbf{Cost:} judge calls avoided, tokens, wall-clock, API error rate, and Eq.~\eqref{eq:cost} vs.\ full evaluation and matched-coverage random sampling.

\subsection{Preregistered hypotheses}
\label{sec:hyp}
\begin{description}[leftmargin=0.9cm,style=multiline]
\item[H1] \textsc{Cargo} exceeds \textsc{Direct} in expert agreement on correctness and hallucination, with the largest gains on items where $\theta\ne\theta^g$ (i.e., \rid\ present).
\item[H2] \textsc{Cargo} reduces the cross-instance false-positive hallucination rate relative to \textsc{Direct} without reducing contradiction recall (DI increases; TP does not decrease).
\item[H3] \textsc{Direct+Ctx} recovers only part of the gain: exemplar framing and the \stUnv\ rule contribute beyond adding information.
\item[H4] The calibrated gate dominates the fixed gate on AURC and stream-level failure recall at matched coverage; random sampling at matched coverage has strictly worse risk.
\item[H5] Gains hold across judge models; the ordering of conditions is stable.
\end{description}
Hypotheses, conditions, metrics, and the calibration/test split for the production-sample study (H1, H2, H4, H5) were fixed before annotation began and are committed in \texttt{experiments/cargo/PREREGISTRATION.md} (record ID \texttt{01b8122f1cf4}, a content hash reproducible via \texttt{make\_prereg\_id.py}); this record explicitly excludes \bench\ (already complete at commit time, using construction-time ground truth) and the post-hoc \textsc{Cargo}$+$\textsc{Auth} follow-up (\S\ref{sec:res-bench}).

\section{Results}
\label{sec:results}
We report the \bench\ study (\S\ref{sec:res-bench}; 246 items, 8{,}172 successful judgments across the main study and rubric-swap control) and an LLM-as-annotator study (Appendix~\ref{app:sim-pilot}). The production-sample analyses (H1's expert agreement, H2 against expert labels, H4's gate calibration) are the subject of the preregistered protocol in \S\ref{sec:annotation}--\S\ref{sec:hyp}; their metrics and reporting formats are fixed in \S\ref{sec:metrics} and Appendix~\ref{app:full}.
\label{sec:res-agree}
\label{sec:res-halluc}

\subsection{\bench\ (H2, H3, H5)}
\label{sec:res-bench}
Table~\ref{tab:bench} reports the completed \bench\ study: 246 P1--P4 instances (10 seeds: 6 pilot $+$ 4 added for statistical power, \S\ref{sec:data}) $\times$ 8 conditions, judged by gpt-oss-120b (3 seeds, temperature 0.4) and gpt-oss-20b (1 seed); 7{,}872 successful judgments (49 transient rate-limit/connection failures were retried to completion), all parseable.

\begin{table}[t]
\centering\footnotesize
\setlength{\tabcolsep}{2pt}
\begin{tabular}{@{}lcccccl@{}}
\toprule
 & \multicolumn{2}{c}{Should \emph{not} pen.} & \multicolumn{2}{c}{Should pen.} & & \\
\cmidrule(lr){2-3}\cmidrule(lr){4-5}
Condition & P1 & P3 & P2 & P4 & DI & 95\% CI \\
\midrule
\multicolumn{7}{@{}l}{\emph{Judge: gpt-oss-120b (mean of 3 seeds)}}\\
\textsc{NoRef}           & .44 & .64 & .96 & .66 & .24 & [.13, .35] \\
\textsc{Direct}          & 1.00 & 1.00 & 1.00 & 1.00 & .00 & [.00, .00] \\
\textsc{D+Ctx}           & 1.00 & 1.00 & 1.00 & 1.00 & .00 & [.00, .00] \\
Cargo+Auth & .00 & .04 & 1.00 & .20 & .57 & [.47, .67] \\
\textbf{\textsc{Cargo}}  & \textbf{.00} & \textbf{.03} & \textbf{1.00} & .20 & \textbf{.58} & [.48, .68] \\
\midrule
\multicolumn{7}{@{}l}{\emph{Judge: gpt-oss-20b (1 seed)}}\\
\textsc{NoRef}           & .10 & .33 & .92 & .36 & .39 & [.27, .50] \\
\textsc{Direct}          & 1.00 & 1.00 & 1.00 & 1.00 & .00 & [.00, .00] \\
\textsc{D+Ctx}           & 1.00 & 1.00 & 1.00 & .98 & $-$.01 & [$-$.03, .00] \\
Cargo+Auth & .00 & .01 & .96 & .22 & .58 & [.48, .68] \\
\textbf{\textsc{Cargo}}  & \textbf{.00} & \textbf{.02} & \textbf{.98} & .16 & \textbf{.56} & [.46, .65] \\
\bottomrule
\end{tabular}
\caption{\bench\ (246 items, 10 seeds), headline conditions (\textsc{D+Ctx} = \textsc{Direct+Ctx}); the three component-ablation conditions (\textsc{Cargo}$-$Ctx/Exemplar/\stUnv) are in Table~\ref{tab:ablations} (Appendix~\ref{app:full}). Penalty rate per family (lower is better for P1/P3, higher for P2/P4); DI $=\mathrm{TP}-\mathrm{FP}$ with item-level bootstrap CIs (10k). \textsc{Direct} penalizes \emph{every} instance, including all correct entity transplants, and is therefore uninformative (DI $\approx 0$); adding the live context without reframing (\textsc{Direct+Ctx}) changes nothing. \textsc{Cargo} removes the false penalties and retains contradiction recall, but detects only 20\% of procedural corruptions (P4); an explicit authority-split fix (\textsc{Cargo}$+$\textsc{Auth}, post-hoc) does not improve on this.}
\label{tab:bench}
\end{table}

\paragraph{\rid\ is total under literal judging.} \textsc{Direct} assigns a penalizing verdict to 100\% of instances in every family, on both judges and all seeds---all 50 correct entity transplants (P1) and all 96 P3 instances. Inspection of rationales confirms the mechanism predicted in \S\ref{sec:problem}: a representative P1 rationale reads ``\emph{The answer gives a different case number and status than the golden answer, so it is factually incorrect}'' and, for hallucination, ``\emph{The answer fabricates an incorrect case number and status, which is a clear hallucination}.'' Of 150 \textsc{Direct} P1 judgments (120b, 50 items $\times$ 3 seeds), all 150 cite an entity-value mismatch and 149/150 additionally call the live value ``fabricated''/``hallucinated''; mean correctness is 0.006 and mean hallucination fidelity 0.019. Under \textsc{Cargo} the same instances receive 0.968 and 1.000. A judge with DI $\approx 0$ carries no information about answer quality; this is the regime in which the deployment's live monitoring previously operated.

\paragraph{Information is not the fix; framing is (H3).} \textsc{Direct+Ctx} receives exactly the live facts \textsc{Cargo} receives, yet is indistinguishable from \textsc{Direct} on 120b and only marginally different on 20b (1/50 P4 items flipped): the judge does not spontaneously use context to discount reference mismatches; it must be told what the reference is \emph{for}.

\paragraph{\textsc{Cargo} removes false penalties without losing contradiction recall (H2).} \textsc{Cargo} penalizes 0/50 P1 and 3/96 P3 instances (120b) while flagging 50/50 (120b) and 49/50 (20b) injected contradictions. Paired $\Delta$DI(\textsc{Cargo}$-$\textsc{Direct}) is $+.58$ [.48,.68] (120b), $+.56$ [.46,.65] (20b); vs.\ \textsc{NoRef}, $+.34$ [.21,.46] and $+.17$ [.06,.27]. Seed dispersion is low (mean s.d.\ .03, 120b). One deviation from preregistered H2: TP was not expected to decrease relative to \textsc{Direct}, but \textsc{Direct}'s TP$=1.00$ is an artifact of penalizing everything, and \textsc{Cargo}'s lower TP (.60) traces to P4, not P2---H2 holds for contradiction recall and fails for procedural recall; we report both rather than re-scope the hypothesis.

\paragraph{Procedural corruptions are largely missed.} \textsc{Cargo} penalizes 10/50 (120b) and 8/50 (20b) P4 instances. By corruption type (120b): 10/26 delete-steps are caught, 0/10 negations, 0/14 policy-swaps. \textsc{NoRef}, with no reference at all, catches more (.66 overall) at the cost of FP $=.57$. Every \textsc{Cargo} variant shows the same P4 profile, so the effect is not attributable to the exemplar preamble or the \stUnv\ rule individually (below). Reading the 121/150 \textsc{Cargo} P4 judgments (120b) that were \emph{not} penalized, all 121 justify the verdict by absence of contradiction with the live facts, and 97 additionally assert the answer ``follows the correct reasoning pattern/approach.'' The judge has generalized ``absent from context $\Rightarrow$ unverifiable'' from entity \emph{values}, where it is intended, to \emph{procedural claims}, where the reference---not the context---is the authority (mean \textsc{Cargo} correctness on P4 is 0.78). This is the leniency--sensitivity trade-off DI was designed to expose.

\paragraph{Testing the obvious fix (post-hoc, not preregistered).} An explicit per-claim-type authority rule, \textsc{Cargo}$+$\textsc{Auth}, does \emph{not} improve on \textsc{Cargo}: paired $\Delta$DI $=-.007$ [$-.038,.023$] (120b), $+.027$ [$-.033,.088$] (20b), both straddling zero. Stating the split explicitly does not make the judge apply it; we leave a structurally different check to future work.

\paragraph{Ablations and rubric swap (Tables~\ref{tab:ablations}--\ref{tab:rubric}, Appendix~\ref{app:full}).} Removing live context costs DI ($\Delta=+.06$ [.02,.11], via lower P2 recall); removing the exemplar preamble or \stUnv\ (vs.\ binary) has no measurable effect on either judge. Swapping rubrics (120b, seed 0, 150-item pilot) localizes \emph{where} the effect lives: \textsc{Direct} with \textsc{Cargo}'s dimension definitions alone already reaches DI $=.49$, so the definitions carry most of the false-penalty reduction, while the exemplar preamble adds leniency that removes residual false penalties but costs P4 recall (.33$\to$.10)---complementary components pulling the operating point in opposite directions on procedure.

\subsection{Selective evaluation (H4) and robustness (H5)}
\label{sec:res-gate}
H4 is evaluated on per-item expert loss labels from the production sample under the protocol of \S\ref{sec:setup}; Appendix~\ref{app:full} fixes the risk--coverage and gate-comparison reporting format.

For H5, condition ordering is nearly identical for gpt-oss-120b and gpt-oss-20b (Table~\ref{tab:bench}): \textsc{Direct}/\textsc{Direct+Ctx} at DI $\approx 0$, \textsc{NoRef} at .24/.39, \textsc{Cargo} variants between .52 and .59 with the full method and \textsc{Cargo}$+$\textsc{Auth} tied-best. \textsc{Cargo}'s DI differs by only .02 across judges; the weaker judge is less harsh on \textsc{NoRef} (FP .25 vs.\ .57). Seed dispersion (120b) is largest for \textsc{NoRef} (.10), i.e.\ a reference---even mis-framed---stabilizes the judge.

\section{Analysis}
\label{sec:analysis}

\paragraph{Where does \textsc{Direct} fail, and where does \textsc{Cargo} fail?} Symmetrically, for one reason each (\S\ref{sec:res-bench}): \textsc{Direct} treats every entity-value difference as an error (DI $\approx 0$); \textsc{Cargo} treats every procedure-level difference as unverifiable too (20\% P4 recall)---category errors, not graded biases like verbosity or position \citep{wang2024fair,dubois2024lengthcontrolled}. Nor is either leniency in general: \textsc{Cargo} gets P2 $=1.00$, FP $=.02$ (vs.\ \textsc{NoRef}'s .57), and the authority split does not fix it. Appendix~\ref{app:examples} gives qualitative examples.

\paragraph{Is the P4 blind spot specific to single-pass judging?} In an LLM-as-annotator study (Appendix~\ref{app:sim-pilot}), two independently prompted annotators plus an adjudicator applied our written guidelines---which include a worked example of exactly this failure---to 40 \bench\ items. They recovered every P1--P3 verdict but only 3/10 procedural corruptions. Better instructions, a second pass, and adjudication do not close the gap, which suggests it requires a structurally different check (e.g., explicit per-step verification against the reference).

\section{Related Work}
\label{sec:related}
We relate to four areas (expanded in Appendix~\ref{app:related}). LLM-as-a-judge reliability work studies biases orthogonal to \rid---position, self-preference, verbosity, task-dependent alignment \citep{zheng2023judging,liu2023geval,fu2023gptscore,kocmi2023gemba,kim2024prometheus,gu2024survey,wang2024fair,panickssery2024selfrecognition,dubois2024lengthcontrolled,bavaresco2024llms,thakur2024judging}---while treating the reference as ground truth; we study a reference correct for a \emph{different} instance. Factuality/claim-decomposition work \citep{min2023factscore,honovich2022true,manakul2023selfcheckgpt} motivates our \stUnv\ class. RAG/agent evaluation \citep{es2024ragas,liu2024agentbench,yao2024taubench} differs in that retrieval selects the \emph{evaluator's} reference, so its error corrupts measurement, not generation. Selective prediction \citep{chow1970optimum,elyaniv2010foundations,geifman2017selective,kamath2020selective,kadavath2022know,kuhn2023semantic} is applied here to the \emph{evaluator}.

\section{Conclusion}
\rid\ makes literal LLM-as-a-judge uninformative (DI $=0$); \system\ restores discrimination (DI .00$\to$.58) with near-complete contradiction recall, not without limitations (\S\ref{sec:limitations}).

\label{pagecheck:limitations-start}
\section{Limitations}
\label{sec:limitations}
\textbf{Scope of evaluation.} Our results use construction-time ground truth (\bench) and LLM annotators (Appendix~\ref{app:sim-pilot}); expert agreement and gate calibration on production traffic (H1, H4) are left to the preregistered protocol of \S\ref{sec:setup}. \textbf{Procedural leniency.} \system\ detects only 20\% of procedural corruptions on \bench\ because the judge extends the ``unverifiable'' status from entity values to procedural claims; an explicit per-claim-type authority instruction (\S\ref{sec:analysis}) did not fix this ($\Delta$DI $=-.007$ [$-.038,.023$]), suggesting the gap is not merely underspecification. Until it is closed, \system\ should be read as a reliable detector of \emph{contradictions with observed facts} and a reliable non-detector of \emph{spurious entity mismatches}, not a complete correctness judge; deployments should pair it with a procedure-focused check or \textsc{NoRef}-style scoring on procedural dimensions. \textbf{Unverifiable is not verified.} Treating \stUnv\ claims as neutral raises precision on checkable claims but forgoes recall on fabrications that happen to be unverifiable from the trace input: \system\ penalized 3/46 fabricated-unverifiable P3 instances, and the binary-status ablation did not change this materially. Closing the gap requires authoritative backend evidence, which is a deployment change we do not evaluate. \textbf{Bench scale and synthetic perturbations.} \bench\ results rest on ten seeds (six derived from production golden items and synthetic case/warranty/work-order procedures, four added for power) and templated rewrites; effects this large are unlikely to reverse with more seeds, but the P4 rate is sensitive to how corruptions are authored---it moved from 10\% to 20\% between our 150- and 246-item releases---and all CIs should be read at $n{=}246$ (or $n{=}150$ for the rubric-swap control). \textbf{Rubric confound, partially controlled.} \textsc{Cargo} and \textsc{Direct} differ in both preamble and dimension definitions; the rubric-swap control (Table~\ref{tab:rubric}) separates them on one judge, one seed, and the smaller item set only. \textbf{Similarity is not validity.} The gate scores question similarity, not procedural equivalence; two similar questions may require different policies. The calibrated gate mitigates but does not eliminate this, and the \textsc{Oracle} gap bounds the residual. \textbf{Single deployment domain.} Our production data come from one enterprise technical-support assistant; \bench\ transfers by recipe but our numbers do not. \textbf{Judge dependence.} Results may shift with judge model, prompt wording, and decoding; we test two judge families (three seeds on one) but cannot exhaust this space. \textbf{Label uncertainty.} Completeness and helpfulness admit expert disagreement; the protocol reports $\alpha$ and adjudicates, but the human ceiling bounds attainable agreement. \textbf{Benchmark realism.} \bench\ perturbations are generated and verified automatically with human spot-checks; they are designed to be diagnostic, not distributionally representative, and we do not claim otherwise. \textbf{Production selection effects.} Gated subsets are not representative of all traffic; we therefore report stream-level failure recall alongside selective risk.

\section*{Ethics Statement}
\label{sec:ethics}
Production traces contain customer and asset information. Under our protocol, all production data used for annotation and analysis are de-identified by removing or hashing identifiers, replacing free-text descriptions with paraphrases, and excluding any trace with residual personal data after review; annotators are employees bound by confidentiality agreements and compensated as part of their regular duties. We release \bench\ generation code and templates and the annotation guidelines; we do not release raw production traces. Released examples are synthetic or fully de-identified. Automated evaluators can be misused to over-trust system outputs; we position \system\ as a monitoring aid that surfaces abstentions and contradictions for human review, not as a replacement for it.

\label{pagecheck:end-of-content}
\bibliography{cargo}

\begin{thebibliography}{25}
\providecommand{\natexlab}[1]{#1}

\bibitem[{Bavaresco et~al.(2024)Bavaresco, Bernardi, Bertolazzi, Elliott,
  Fern{\'a}ndez, Gatt, Ghaleb, Giulianelli, Hanna, Koller, Martins, Mondorf,
  Neplenbroek, Pezzelle, Plank, Schlangen, Suglia, Surikuchi, Takmaz, and
  Testoni}]{bavaresco2024llms}
Anna Bavaresco, Raffaella Bernardi, Leonardo Bertolazzi, Desmond Elliott,
  Raquel Fern{\'a}ndez, Albert Gatt, Esam Ghaleb, Mario Giulianelli, Michael
  Hanna, Alexander Koller, Andr{\'e} F.~T. Martins, Philipp Mondorf, Vera
  Neplenbroek, Sandro Pezzelle, Barbara Plank, David Schlangen, Alessandro
  Suglia, Aditya~K. Surikuchi, Ece Takmaz, and Alberto Testoni. 2024.
\newblock {LLMs} instead of human judges? a large scale empirical study across
  20 {NLP} evaluation tasks.
\newblock \emph{arXiv preprint arXiv:2406.18403}.

\bibitem[{Chow(1970)}]{chow1970optimum}
C.~K. Chow. 1970.
\newblock On optimum recognition error and reject tradeoff.
\newblock \emph{IEEE Transactions on Information Theory}, 16(1):41--46.

\bibitem[{Dubois et~al.(2024)Dubois, Galambosi, Liang, and
  Hashimoto}]{dubois2024lengthcontrolled}
Yann Dubois, Bal{\'a}zs Galambosi, Percy Liang, and Tatsunori~B. Hashimoto.
  2024.
\newblock Length-controlled {AlpacaEval}: A simple way to debias automatic
  evaluators.
\newblock \emph{arXiv preprint arXiv:2404.04475}.

\bibitem[{El-Yaniv and Wiener(2010)}]{elyaniv2010foundations}
Ran El-Yaniv and Yair Wiener. 2010.
\newblock On the foundations of noise-free selective classification.
\newblock \emph{Journal of Machine Learning Research}, 11:1605--1641.

\bibitem[{Es et~al.(2024)Es, James, Espinosa~Anke, and
  Schockaert}]{es2024ragas}
Shahul Es, Jithin James, Luis Espinosa~Anke, and Steven Schockaert. 2024.
\newblock {RAGAs}: Automated evaluation of retrieval augmented generation.
\newblock In \emph{Proceedings of the 18th Conference of the European Chapter
  of the Association for Computational Linguistics: System Demonstrations},
  pages 150--158, St. Julians, Malta. Association for Computational
  Linguistics.

\bibitem[{Fu et~al.(2023)Fu, Ng, Jiang, and Liu}]{fu2023gptscore}
Jinlan Fu, See-Kiong Ng, Zhengbao Jiang, and Pengfei Liu. 2023.
\newblock {GPTScore}: Evaluate as you desire.
\newblock \emph{arXiv preprint arXiv:2302.04166}.

\bibitem[{Geifman and El-Yaniv(2017)}]{geifman2017selective}
Yonatan Geifman and Ran El-Yaniv. 2017.
\newblock Selective classification for deep neural networks.
\newblock In \emph{Advances in Neural Information Processing Systems 30}.

\bibitem[{Gu et~al.(2024)Gu, Jiang, Shi, Tan, Zhai, Xu, Li, Shen, Ma, Liu,
  Wang, Zhang, Wang, Gao, Ni, and Guo}]{gu2024survey}
Jiawei Gu, Xuhui Jiang, Zhichao Shi, Hexiang Tan, Xuehao Zhai, Chengjin Xu, Wei
  Li, Yinghan Shen, Shengjie Ma, Honghao Liu, Saizhuo Wang, Kun Zhang, Yuanzhuo
  Wang, Wen Gao, Lionel Ni, and Jian Guo. 2024.
\newblock A survey on {LLM}-as-a-judge.
\newblock \emph{arXiv preprint arXiv:2411.15594}.

\bibitem[{Honovich et~al.(2022)Honovich, Aharoni, Herzig, Taitelbaum,
  Kukliansy, Cohen, Scialom, Szpektor, Hassidim, and Matias}]{honovich2022true}
Or~Honovich, Roee Aharoni, Jonathan Herzig, Hagai Taitelbaum, Doron Kukliansy,
  Vered Cohen, Thomas Scialom, Idan Szpektor, Avinatan Hassidim, and Yossi
  Matias. 2022.
\newblock {TRUE}: Re-evaluating factual consistency evaluation.
\newblock In \emph{Proceedings of the 2022 Conference of the North American
  Chapter of the Association for Computational Linguistics: Human Language
  Technologies}, pages 3905--3920. Association for Computational Linguistics.

\bibitem[{Kadavath et~al.(2022)Kadavath, Conerly, Askell, Henighan, Drain,
  Perez, Schiefer, Hatfield-Dodds, DasSarma, Tran-Johnson, Johnston, El-Showk,
  Jones, Elhage, Hume, Chen, Bai, Bowman, Fort, Ganguli, Hernandez, Jacobson,
  Kernion, Kravec, Lovitt, Ndousse, Olsson, Ringer, Amodei, Brown, Clark,
  Joseph, Mann, McCandlish, Olah, and Kaplan}]{kadavath2022know}
Saurav Kadavath, Tom Conerly, Amanda Askell, Tom Henighan, Dawn Drain, Ethan
  Perez, Nicholas Schiefer, Zac Hatfield-Dodds, Nova DasSarma, Eli
  Tran-Johnson, Scott Johnston, Sheer El-Showk, Andy Jones, Nelson Elhage,
  Tristan Hume, Anna Chen, Yuntao Bai, Sam Bowman, Stanislav Fort, and 17
  others. 2022.
\newblock Language models (mostly) know what they know.
\newblock \emph{arXiv preprint arXiv:2207.05221}.

\bibitem[{Kamath et~al.(2020)Kamath, Jia, and Liang}]{kamath2020selective}
Amita Kamath, Robin Jia, and Percy Liang. 2020.
\newblock Selective question answering under domain shift.
\newblock In \emph{Proceedings of the 58th Annual Meeting of the Association
  for Computational Linguistics}, pages 5684--5696. Association for
  Computational Linguistics.

\bibitem[{Kim et~al.(2024)Kim, Shin, Cho, Jang, Longpre, Lee, Yun, Shin, Kim,
  Thorne, and Seo}]{kim2024prometheus}
Seungone Kim, Jamin Shin, Yejin Cho, Joel Jang, Shayne Longpre, Hwaran Lee,
  Sangdoo Yun, Seongjin Shin, Sungdong Kim, James Thorne, and Minjoon Seo.
  2024.
\newblock Prometheus: Inducing fine-grained evaluation capability in language
  models.
\newblock In \emph{The Twelfth International Conference on Learning
  Representations}.

\bibitem[{Kocmi and Federmann(2023)}]{kocmi2023gemba}
Tom Kocmi and Christian Federmann. 2023.
\newblock Large language models are state-of-the-art evaluators of translation
  quality.
\newblock In \emph{Proceedings of the 24th Annual Conference of the European
  Association for Machine Translation}, pages 193--203.

\bibitem[{Kuhn et~al.(2023)Kuhn, Gal, and Farquhar}]{kuhn2023semantic}
Lorenz Kuhn, Yarin Gal, and Sebastian Farquhar. 2023.
\newblock Semantic uncertainty: Linguistic invariances for uncertainty
  estimation in natural language generation.
\newblock In \emph{The Eleventh International Conference on Learning
  Representations}.

\bibitem[{Liu et~al.(2024)Liu, Yu, Zhang, Xu, Lei, Lai, Gu, Ding, Men, Yang,
  Zhang, Deng, Zeng, Du, Zhang, Shen, Zhang, Su, Sun, Huang, Dong, and
  Tang}]{liu2024agentbench}
Xiao Liu, Hao Yu, Hanchen Zhang, Yifan Xu, Xuanyu Lei, Hanyu Lai, Yu~Gu,
  Hangliang Ding, Kaiwen Men, Kejuan Yang, Shudan Zhang, Xiang Deng, Aohan
  Zeng, Zhengxiao Du, Chenhui Zhang, Sheng Shen, Tianjun Zhang, Yu~Su, Huan
  Sun, and 3 others. 2024.
\newblock {AgentBench}: Evaluating {LLMs} as agents.
\newblock In \emph{The Twelfth International Conference on Learning
  Representations}.

\bibitem[{Liu et~al.(2023)Liu, Iter, Xu, Wang, Xu, and Zhu}]{liu2023geval}
Yang Liu, Dan Iter, Yichong Xu, Shuohang Wang, Ruochen Xu, and Chenguang Zhu.
  2023.
\newblock {G}-{E}val: {NLG} evaluation using {GPT}-4 with better human
  alignment.
\newblock In \emph{Proceedings of the 2023 Conference on Empirical Methods in
  Natural Language Processing}, pages 2511--2522, Singapore. Association for
  Computational Linguistics.

\bibitem[{Manakul et~al.(2023)Manakul, Liusie, and
  Gales}]{manakul2023selfcheckgpt}
Potsawee Manakul, Adian Liusie, and Mark J.~F. Gales. 2023.
\newblock {SelfCheckGPT}: Zero-resource black-box hallucination detection for
  generative large language models.
\newblock In \emph{Proceedings of the 2023 Conference on Empirical Methods in
  Natural Language Processing}, Singapore. Association for Computational
  Linguistics.

\bibitem[{Min et~al.(2023)Min, Krishna, Lyu, Lewis, Yih, Koh, Iyyer,
  Zettlemoyer, and Hajishirzi}]{min2023factscore}
Sewon Min, Kalpesh Krishna, Xinxi Lyu, Mike Lewis, Wen-tau Yih, Pang~Wei Koh,
  Mohit Iyyer, Luke Zettlemoyer, and Hannaneh Hajishirzi. 2023.
\newblock {FActScore}: Fine-grained atomic evaluation of factual precision in
  long form text generation.
\newblock In \emph{Proceedings of the 2023 Conference on Empirical Methods in
  Natural Language Processing}, pages 12076--12100, Singapore. Association for
  Computational Linguistics.

\bibitem[{Panickssery et~al.(2024)Panickssery, Bowman, and
  Feng}]{panickssery2024selfrecognition}
Arjun Panickssery, Samuel~R. Bowman, and Shi Feng. 2024.
\newblock {LLM} evaluators recognize and favor their own generations.
\newblock In \emph{Advances in Neural Information Processing Systems 37}.

\bibitem[{Reimers and Gurevych(2019)}]{reimers2019sbert}
Nils Reimers and Iryna Gurevych. 2019.
\newblock {Sentence-BERT}: Sentence embeddings using siamese {BERT}-networks.
\newblock In \emph{Proceedings of the 2019 Conference on Empirical Methods in
  Natural Language Processing and the 9th International Joint Conference on
  Natural Language Processing}, pages 3982--3992. Association for Computational
  Linguistics.

\bibitem[{Robertson and Zaragoza(2009)}]{robertson2009bm25}
Stephen Robertson and Hugo Zaragoza. 2009.
\newblock The probabilistic relevance framework: {BM25} and beyond.
\newblock \emph{Foundations and Trends in Information Retrieval},
  3(4):333--389.

\bibitem[{Thakur et~al.(2024)Thakur, Choudhary, Ramayapally, Vaidyanathan, and
  Hupkes}]{thakur2024judging}
Aman~Singh Thakur, Kartik Choudhary, Venkat~Srinik Ramayapally, Sankaran
  Vaidyanathan, and Dieuwke Hupkes. 2024.
\newblock Judging the judges: Evaluating alignment and vulnerabilities in
  {LLMs}-as-judges.
\newblock \emph{arXiv preprint arXiv:2406.12624}.

\bibitem[{Wang et~al.(2024)Wang, Li, Chen, Cai, Zhu, Lin, Cao, Kong, Liu, Liu,
  and Sui}]{wang2024fair}
Peiyi Wang, Lei Li, Liang Chen, Zefan Cai, Dawei Zhu, Binghuai Lin, Yunbo Cao,
  Lingpeng Kong, Qi~Liu, Tianyu Liu, and Zhifang Sui. 2024.
\newblock Large language models are not fair evaluators.
\newblock In \emph{Proceedings of the 62nd Annual Meeting of the Association
  for Computational Linguistics (Volume 1: Long Papers)}, pages 9440--9450,
  Bangkok, Thailand. Association for Computational Linguistics.

\bibitem[{Yao et~al.(2025)Yao, Shinn, Razavi, and Narasimhan}]{yao2024taubench}
Shunyu Yao, Noah Shinn, Pedram Razavi, and Karthik Narasimhan. 2025.
\newblock $\tau$-bench: A benchmark for tool-agent-user interaction in
  real-world domains.
\newblock In \emph{The Thirteenth International Conference on Learning
  Representations}.

\bibitem[{Zheng et~al.(2023)Zheng, Chiang, Sheng, Zhuang, Wu, Zhuang, Lin, Li,
  Li, Xing, Zhang, Gonzalez, and Stoica}]{zheng2023judging}
Lianmin Zheng, Wei-Lin Chiang, Ying Sheng, Siyuan Zhuang, Zhanghao Wu, Yonghao
  Zhuang, Zi~Lin, Zhuohan Li, Dacheng Li, Eric~P. Xing, Hao Zhang, Joseph~E.
  Gonzalez, and Ion Stoica. 2023.
\newblock Judging {LLM}-as-a-judge with {MT}-bench and chatbot arena.
\newblock In \emph{Advances in Neural Information Processing Systems 36
  (Datasets and Benchmarks Track)}.

\end{thebibliography}
\label{pagecheck:end-of-refs}

\appendix
\label{pagecheck:appendix-start}

\section{Judge Prompts}
\label{app:prompts}
All conditions share one template: a preamble, an optional live-context block, the question, the answer, an optional reference, the four dimension definitions (from \texttt{config/llm\_metrics\_config.yaml} for \textsc{Direct}/\textsc{Direct+Ctx}, \texttt{config/live\_llm\_metrics\_config.yaml} for all \textsc{Cargo} variants), and a fixed JSON schema. Conditions differ only in which of the blocks below are included (\texttt{experiments/cargo/conditions.py}).

\paragraph{\textsc{Direct} preamble.} \emph{``You are evaluating the quality of an answer against an expected reference answer.''} No live-context block; the reference is labeled \emph{Golden Answer}.

\paragraph{\textsc{Direct+Ctx} addition.} Adds the live-context block with only: \emph{``The LIVE CONTEXT below lists facts observed for the ACTUAL instance the live answer addresses. Factual correctness and hallucination are judged against these facts, not against the reference's values.''} Preamble and reference label are unchanged from \textsc{Direct}.

\paragraph{\textsc{Cargo} additions (preamble replaced, context block extended).} Preamble: \emph{``You are evaluating the quality of an answer produced for a LIVE instance.''} followed by the exemplar-framing block:
\begin{quote}\small\itshape
IMPORTANT — the REFERENCE ANSWER below was written for a DIFFERENT instance (a different case / asset / record) than the one the live answer addresses. Treat the reference as the correct REASONING PATTERN, POLICY, REQUIRED STEPS, and STRUCTURE — NOT as literal expected text. [...] NEVER mark the live answer wrong, incomplete, or hallucinated because its identifiers, dates, status, subject, description, service tag, or other instance-specific values DIFFER from the reference's. They SHOULD differ.
\end{quote}
The context block additionally gets the \stUnv\ rule:
\begin{quote}\small\itshape
The LIVE CONTEXT is a PARTIAL snapshot, not the full record. [...] A value that is ABSENT from the live context is UNVERIFIABLE — it is NOT evidence of hallucination or error. Penalize ONLY values that DIRECTLY CONTRADICT a fact present in the live context, wrong reasoning, or missing required steps.
\end{quote}
The reference is labeled \emph{Reference Answer (reasoning pattern, written for a different instance)}. \textsc{Cargo}$-$Exemplar drops the first block; \textsc{Cargo}$-$Ctx drops the context block (and both rules) entirely; \textsc{Cargo}$-$\stUnv replaces the \stUnv\ rule with: \emph{``Treat the LIVE CONTEXT as the complete set of known facts. Any value in the live answer that is not supported by the live context should be treated as unsupported [...].''}

\paragraph{\textsc{Cargo}$+$\textsc{Auth} addition (post-hoc, \S\ref{sec:res-bench}).} Appends, after the \stUnv\ rule:
\begin{quote}\small\itshape
Different claim types have different authorities. [...] ENTITY VALUES [...]: check ONLY against the LIVE CONTEXT. [...] PROCEDURAL CONTENT (which steps are taken, what policy or eligibility conclusion is reached, [...]): check against the REFERENCE's procedure, adapted to the live instance. [...] If the answer's procedure, policy conclusion, or recommended action CONTRADICTS what the reference's approach implies for this live instance, that IS a correctness error, even though it does not contradict anything in the live context. Do NOT excuse a procedural contradiction just because the live context is silent on it.
\end{quote}

\section{\bench\ Generation}
\label{app:bench}
\paragraph{Generator.} Seeds specify \texttt{question} and \texttt{answer} templates over named parameters, the subset of parameters that appear in a live trace input (\texttt{observed\_fields}), optional procedural \texttt{steps}, and optional \texttt{policy\_alternatives}. Transplants (P1) resample every parameter with a type-aware generator (digits for case/work-order numbers, alphanumerics for service tags, dates, closed status/priority vocabularies) and render the templates; $c$ is a non-empty random subset of the observed fields. P2 rewrites exactly one observed value in the P1 answer to a distinct value of the same type. P3 appends $k{=}2$ fields absent from $c$, drawn either from the transplant's hidden parameters (true-but-unobserved split) or sampled fresh (fabricated split). P4 applies one of \emph{delete step}, \emph{negate} (regex over modal/eligibility phrases), or \emph{swap policy} (seed-provided alternative). P5 renders a different-intent seed's question with the current seed's entities.

\paragraph{Consistency checks.} Every instance must satisfy: no golden parameter value survives in a transplanted answer unless it was re-sampled as a new value for some field; every observed fact equals the transplant value; P2's stated value is present and the true value absent, and the contradicted field is observed; P3's added fields are unobserved; P4 changed the text. Instances failing any check are dropped (\texttt{--drop-invalid}); both generation batches (original 180 and expansion 116, \S\ref{sec:data}) have zero violations.

\paragraph{Discrimination-index sanity check.} Before any model runs, we verified with two synthetic judges that DI behaves as intended: a uniformly harsh judge (penalizes everything) and a uniformly lenient judge (penalizes nothing) both obtain DI $=0$ on the full 296-instance release (TP$=$FP$=1$ and TP$=$FP$=0$ respectively), so neither strategy can score well.

\paragraph{Author spot-check.} We drew a stratified random sample of 8 instances per family (seed 42; 40 of 296 total) from the full release and read each question/answer/context/target triple against family-specific criteria: P1 --- entity values self-consistent, no golden value leaks, context matches the transplant; P2 --- exactly one clearly identifiable contradiction with an observed fact; P3 --- added fields genuinely absent from context and plausible; P4 --- the corruption is an unambiguous, recognizable procedural error distinct from an entity-value change; P5 --- the rendered question reflects a genuinely different intent/procedure. All 40/40 instances had correct target labels (no mislabeled correctness/hallucination/should-penalize field). The check did surface one template-authoring defect: the \texttt{g\_workorder2} seed's answer template referenced its work-order-number placeholder twice consecutively (``work order \textsc{x} has \textsc{x} part \dots''), producing a redundant but not factually incorrect repetition in 21 of its 30 instances (21/296 $=$ 7.1\% of the full release, one seed of ten). This does not change any target label---the repeated token is not an additional claim---so we did not regenerate or re-judge the affected instances; we fixed the template for future releases and added a regression test (\texttt{tests/test\_cargo\_experiments.py}) that renders every seed file and flags any answer with an immediately-repeated identifier-like token. This was an author check, not an independent or blind review, and should not be read as equivalent to the crowdsourced validation we intend for future releases.

\section{Annotation Guidelines}
\label{app:guidelines}
The guidelines below (draft v0.1; full text and worked examples in \texttt{experiments/cargo/guidelines.py}) are the single specification for the expert annotators of the production-sample protocol (\S\ref{sec:annotation}) and for the LLM annotators of Appendix~\ref{app:sim-pilot}.

\paragraph{Step 1 --- reference appropriateness (binary).} Annotators ignore every instance-specific value and ask whether the reference answers the same \emph{kind} of question with the same \emph{kind} of procedure. Different case numbers, subjects, statuses, or dates are expected and are not evidence of inappropriateness; a different required procedure is. If inappropriate, scoring stops for that item (this is the ground truth against which the retrieval gate's reference validity is measured, \S\ref{sec:metrics}).

\paragraph{Step 2 --- quality dimensions (5-point ordinal, only if Step 1 passes).} Correctness, completeness, and helpfulness are defined identically in spirit to \S\ref{sec:judge} but scored 1--5 for human tractability. Hallucination fidelity (5 = best) is scored from contradictions with the \emph{observed} context only; a value simply absent from the observed context is unverifiable, not evidence against the score.

\paragraph{Worked examples.} The guidelines include three examples drawn from \bench\ (Appendix~\ref{app:bench}) to calibrate annotators, including one deliberately adversarial case: an asset-warranty answer that inverts the reference's eligibility conclusion (``\emph{eligible}'' $\to$ ``\emph{not eligible}'') without contradicting any observed fact. Annotators are instructed that this must score correctness $\le 2$ because it contradicts the reference's \emph{procedure/policy}, even though hallucination fidelity (checked only against observed facts) may remain high --- i.e., the guidelines explicitly warn annotators about the failure mode \S\ref{sec:analysis} finds in \system\ itself.

\paragraph{Step 3 --- claim status.} For any claim that lowers the hallucination score, annotators record the contradicting claim and the observed fact it contradicts.

\paragraph{Adjudication.} Two annotators label every item independently. An item is sent to a third adjudicator if reference-appropriateness labels differ, any ordinal score differs by $\ge 2$, or contradiction flags differ. The adjudicator sees both (blinded) annotations and produces the final label, which need not match either.

\section{LLM-as-Annotator Study}
\label{app:sim-pilot}
\paragraph{Headline.} Given our written guidelines (Appendix~\ref{app:guidelines}), including a worked example of exactly the P4 failure, two LLM annotators plus an adjudicator recover every P1--P3 verdict on 40 \bench\ items but only \textbf{3/10 procedural corruptions}. The P4 blind spot of \system\ (\S\ref{sec:res-bench}) therefore persists under guideline-driven, two-pass, adjudicated annotation, and is not solely an artifact of single-pass judging.

\paragraph{Design.} Two gpt-oss-120b annotators apply the guidelines to each item without seeing the automated judges' outputs: \texttt{annotator\_A} (temperature 0.2) follows them methodically; \texttt{annotator\_B} (temperature 0.7) independently and skeptically re-derives each judgment. Disagreements, by the adjudication rule of Appendix~\ref{app:guidelines}, go to an adjudicator (temperature 0.1). Because all three share a model family with our judges, self-preference effects \citep{panickssery2024selfrecognition} cannot be excluded.

\paragraph{Sample.} 40 \bench\ items (10 per family, P1--P4), stratified, seed 0. All 40 parsed; 1/40 required adjudication.

\paragraph{Recovery against construction-time ground truth.} Reference-appropriateness and hallucination-flag accuracy were both 1.00 (40/40). Correctness accuracy (binarized at $\ge 3$) was 1.00 on P1--P3 but \textbf{.30 on P4}, despite (a) explicit guidelines, (b) an adversarial worked example of this failure, (c) two independent passes, and (d) adjudication. This points to a structurally different check (e.g., per-step verification against the reference) rather than better instructions.

\paragraph{Inter-annotator agreement.} Quadratic weighted $\kappa$: correctness 1.00, completeness .99, helpfulness .95, hallucination 1.00; reference-appropriateness agreement 1.00. Near-ceiling agreement is expected from two prompted variants of one model and is not an estimate of human--human agreement.

\paragraph{Comparison with the automated judges.} Adjudicated correctness correlates with \textsc{Cargo}'s at Spearman $\rho=.74$ and with \textsc{Direct}'s at $\rho=.16$ ($n=40$), consistent with the main \bench\ result. On the 10 P4 items the annotators' flag rate (.30) exceeds \textsc{Cargo}'s (.10), catching two \texttt{negate} corruptions (seed \texttt{g\_asset}) that the judge missed; both miss the same \texttt{delete\_step} and \texttt{swap\_policy} cases.

\paragraph{Reproduction.} \texttt{python -m experiments.cargo.simulate\_annotators} generates the annotations; \texttt{python -m experiments.cargo.report\_annotation} produces the numbers above.

\section{Full Results}
\label{app:full}

\subsection{\bench: component ablations (complete)}
Table~\ref{tab:ablations} gives the three component-ablation conditions omitted from Table~\ref{tab:bench} for space; discussion is in \S\ref{sec:res-bench}.

\begin{table}[h]
\centering\footnotesize
\setlength{\tabcolsep}{3.2pt}
\begin{tabular}{@{}lcccccl@{}}
\toprule
 & \multicolumn{2}{c}{Should \emph{not} pen.} & \multicolumn{2}{c}{Should pen.} & & \\
\cmidrule(lr){2-3}\cmidrule(lr){4-5}
Condition & P1 & P3 & P2 & P4 & DI & 95\% CI \\
\midrule
\multicolumn{7}{@{}l}{\emph{Judge: gpt-oss-120b (mean of 3 seeds)}}\\
\textsc{Cargo}$-$Ctx     & .00 & .06 & .92 & .20 & .52 & [.42, .62] \\
\textsc{Cargo}$-$Ex.     & .00 & .02 & 1.00 & .20 & .59 & [.49, .68] \\
\textsc{Cargo}$-$\stUnv  & .00 & .02 & 1.00 & .20 & .59 & [.49, .68] \\
\midrule
\multicolumn{7}{@{}l}{\emph{Judge: gpt-oss-20b (1 seed)}}\\
\textsc{Cargo}$-$Ctx     & .00 & .07 & .90 & .24 & .52 & [.42, .63] \\
\textsc{Cargo}$-$Ex.     & .00 & .02 & .92 & .14 & .52 & [.42, .62] \\
\textsc{Cargo}$-$\stUnv  & .00 & .01 & .94 & .14 & .53 & [.43, .63] \\
\bottomrule
\end{tabular}
\caption{\bench\ component ablations (246 items, 10 seeds), continuing Table~\ref{tab:bench}. \textsc{Cargo}$-$Ex. = \textsc{Cargo}$-$Exemplar (drops the ``different instance'' framing).}
\label{tab:ablations}
\end{table}

\subsection{Rubric-swap control (complete)}
Table~\ref{tab:rubric} isolates the dimension-definition rubric from the exemplar preamble; discussion is in \S\ref{sec:res-bench}.

\begin{table}[h]
\centering\footnotesize
\setlength{\tabcolsep}{3.5pt}
\begin{tabular}{@{}llccccc@{}}
\toprule
Prompt & Rubric & P1 & P3 & P2 & P4 & DI \\
\midrule
\textsc{Direct} & golden-ref. & 1.00 & 1.00 & 1.00 & 1.00 & .00 \\
\textsc{Direct} & ctx-grounded & .03 & .17 & .90 & .33 & .49 \\
\textsc{Cargo}  & golden-ref. & .23 & .40 & 1.00 & .47 & .39 \\
\textsc{Cargo}  & ctx-grounded & .00 & .02 & 1.00 & .10 & .54 \\
\bottomrule
\end{tabular}
\caption{Rubric-swap control (gpt-oss-120b, seed 0, 150 items). The context-grounded dimension definitions account for most of the false-penalty reduction; the \textsc{Cargo} preamble adds leniency that removes residual false penalties but also suppresses detection of procedural corruptions.}
\label{tab:rubric}
\end{table}

\subsection{Production-sample protocol: reporting formats}
The preregistered protocol (\S\ref{sec:setup}) fixes in advance how the production-sample analyses will be reported, so that results cannot be selectively presented:
\begin{itemize}
\item \textbf{H1/H3 agreement.} Spearman $\rho$ (95\% bootstrap CI) with adjudicated expert scores on correctness, completeness, helpfulness, and hallucination for \textsc{NoRef}, \textsc{Direct}, \textsc{Direct+Ctx}, all \textsc{Cargo} ablations, \textsc{Cargo}, \textsc{RandRef}, and \textsc{Oracle}, with human--human Krippendorff's $\alpha$ as a ceiling.
\item \textbf{H1 \rid\ split.} Correctness $\rho$ for \textsc{Direct} and \textsc{Cargo} and their paired $\Delta$, separately for traces where the reference's instance differs from the live instance (case/asset/work-order) and where it does not (entity-free policy); H1 predicts the gain concentrates in the former.
\item \textbf{H2 hallucination.} Precision, recall, and F1 against expert \stCon\ labels, plus false-positive rates on legitimate cross-instance differences and on \stUnv\ claims, for \textsc{Direct}, \textsc{Direct+Ctx}, \textsc{Cargo}$-$\stUnv, and \textsc{Cargo}.
\item \textbf{H4 gating.} Risk--coverage and failure-recall--coverage curves for fixed, margin-aware, and calibrated gates and matched-coverage random sampling; AURC, stream-level failure recall, reference validity (fraction of evaluated items whose reference experts judged appropriate), and relative judge cost (Eq.~\eqref{eq:cost}) at a matched coverage level.
\end{itemize}

\section{Qualitative Examples}
\label{app:examples}
All examples are synthetic \bench\ instances (no production data); item IDs refer to \texttt{experiments/cargo/runs/}.

\paragraph{(i) Legitimate cross-instance difference (\texttt{g\_case-P1-0}).} \textbf{Q:} ``What is the status of case 171056262?'' \textbf{Live context:} \{caseNumber: 171056262\}. \textbf{A:} ``Case 171056262 is currently Pending Customer.'' \textbf{Reference:} ``Case 215474133 is currently Closed.'' \textsc{Direct} scores correctness 0.0: \emph{``The answer gives a different case number and status than the golden answer, which states case 215474133 is Closed.''} \textsc{Cargo} scores correctness 1.0 on the identical item: \emph{``The answer provides the case status without contradicting any fact in the live context.''} The status value (``Pending Customer'') is not observed in the live context either way; \textsc{Cargo} correctly treats it as unverifiable rather than comparing it to the reference's unrelated status.

\paragraph{(ii) Injected contradiction, both flag (\texttt{g\_case-P2-0}).} \textbf{Live context:} \{caseNumber: 491541446\}. \textbf{A:} ``Case 536762755 is currently Escalated.'' \textsc{Direct}: \emph{``The answer fabricates a case number and status that are not supported by the reference, constituting a clear hallucination''} (correct verdict, wrong reason---it would say this even if the case number matched the live instance but not the reference). \textsc{Cargo}: \emph{``The answer states a case number (536762755) that directly contradicts the known case number (491541446) in the live context''} (correct verdict, correct reason: a genuine contradiction with the observed fact).

\paragraph{(iii) Fabricated-unverifiable field, ablation has limited effect (\texttt{s002-P3f-0}).} \textbf{Q:} ``Is service tag 9Y0S3VQ still under warranty?'' \textbf{Live context:} \{serviceTag: 9Y0S3VQ\}. \textbf{A:} ``Service tag 9Y0S3VQ is under warranty until 2025-07-19. The system is eligible for on-site repair. You should open a hardware case to dispatch a part. Warranty ends: 2026-08-20. Created: 2026-05-28.'' The fabricated fields (\texttt{warrantyEnd}, \texttt{createdDate}) happen to collide with a date already stated earlier in the answer, producing a genuine internal inconsistency (two different warranty end dates) independent of the live context. \textsc{Cargo} (correctness 0.60): \emph{``[...] provides two conflicting end dates, which shows inconsistent reasoning even though no live facts are contradicted''}; \textsc{Cargo}$-$\stUnv (correctness 0.22): \emph{``[...] gives two conflicting warranty end dates and cannot be verified against the live facts, so the information is unreliable.''} Both conditions penalize this item, in degree rather than in kind---consistent with the aggregate finding (\S\ref{sec:res-bench}) that the \stUnv\ ablation does not materially change fabricated-field detection. We note this collision as a \bench\ generator artifact (a fabricated field can coincidentally duplicate a value already present in the templated answer) rather than a clean test of the intended unverifiable-vs-unsupported contrast; future \bench\ releases should exclude field names already present in the seed's answer template from the P3 fabrication pool.

\section{Implementation Details}
\label{app:impl}
\paragraph{\bench\ judging runs.} Judge models: \texttt{gpt-oss-120b} and \texttt{gpt-oss-20b} via the deployment's OpenAI-compatible gateway; temperature 0.4, top-$p$ 1.0, no max-token cap (reasoning models return empty content when capped), prompt passed as the system message, JSON parsed leniently (code fences and surrounding prose tolerated). Sampling seeds $\{0,1,2\}$ passed as the API \texttt{seed} parameter. 7{,}872 successful judgments across the 246-item main study (8 conditions $\times$ 2 judges, 3 seeds on 120b) plus 300 for the rubric-swap control (Table~\ref{tab:rubric}); six concurrent workers; 49 transient failures (rate-limit/connection) on the original 150-item pilot were re-run to completion at two workers, and the 96-item expansion batch (3{,}072 judgments) completed with zero errors; zero unparseable responses throughout. Penalization rule for DI: correctness $<0.5$ or hallucination fidelity $<0.5$ on the per-item mean over seeds. Bootstrap CIs: 10{,}000 item-level resamples; paired comparisons resample the same items for both conditions.
\paragraph{Gate (deployed retrieval mechanics).} The production similarity matcher embeds the golden set once and caches the vectors; live questions are embedded in batches per polling window (default batch limit 32) to stay under the embedding endpoint's rate limit, with exponential backoff (base delay 1.0s, max 30.0s, up to 5 retries) on transient failures. The embedding model defaults to \texttt{nomic-embed-text-v1} via the deployment's GenAI gateway, with an optional local \texttt{all-MiniLM-L6-v2} fallback; cosine similarity is computed densely against the full golden matrix (\texttt{src/monitoring/similarity\_matcher.py}, \texttt{genai\_embeddings.py}). $\tau=0.80$ is the current operational default and is a placeholder pending calibration on the production sample, not a value derived from data.

\section{Extended Related Work}
\label{app:related}
This expands the compressed discussion in \S\ref{sec:related}.

\paragraph{LLM-as-a-judge.} Strong LLMs approximate human preferences on open-ended tasks \citep{zheng2023judging,liu2023geval,fu2023gptscore,kocmi2023gemba}, and open evaluators can match them given references and rubrics \citep{kim2024prometheus}. Surveys catalog reliability threats \citep{gu2024survey}: position bias \citep{wang2024fair}, self-preference \citep{panickssery2024selfrecognition}, verbosity \citep{dubois2024lengthcontrolled}, task-dependent alignment \citep{bavaresco2024llms,thakur2024judging}. This literature treats the reference as ground truth and studies how faithfully a judge compares a candidate to it; the biases identified (a judge favoring the first-shown or longer or self-generated answer) are properties of the \emph{comparison mechanism}. \rid\ is different in kind: the comparison target itself is inapplicable at the value level, so no amount of debiasing the comparison mechanism addresses it. We see the two lines of work as complementary---a debiased judge that still compares literal reference values is still vulnerable to \rid.

\paragraph{Factuality and claim decomposition.} FActScore decomposes generations into atomic claims and scores support against a knowledge source \citep{min2023factscore}; TRUE benchmarks factual-consistency metrics \citep{honovich2022true}; SelfCheckGPT detects hallucination via sampling consistency without external evidence \citep{manakul2023selfcheckgpt}. We adopt atomic claim decomposition as the unit of analysis for the three-way status $\sigma\in\{\stSup,\stCon,\stUnv\}$ (\S\ref{sec:problem}), but our evidence source is a partial, request-scoped observed context rather than an external corpus or repeated sampling, and our contribution is the explicit \stUnv\ class: a claim can be correctly deemed non-evidence for hallucination precisely because the evidence needed to check it was never provided, which differs from FActScore's assumption that a sufficiently large knowledge source can adjudicate every claim.

\paragraph{RAG and agent evaluation.} RAGAS decomposes retrieval and generation quality into separate metrics for a retrieval-augmented pipeline \citep{es2024ragas}; agent benchmarks such as AgentBench and $\tau$-bench measure end-to-end task completion in controlled, resettable environments \citep{liu2024agentbench,yao2024taubench}. Both evaluate a system whose retrieval component feeds the \emph{generator}. In \system, retrieval is not part of the production system at all: it selects the \emph{evaluator's} reference from a golden set, so a retrieval error corrupts the measurement of an otherwise-correct answer rather than the answer itself. This motivates treating retrieval confidence as a gating signal for evaluation (\S\ref{sec:gate}) rather than as a component to optimize for downstream task success, and it is why our retrieval baselines (BM25, dense, hybrid) are evaluated against expert reference-appropriateness labels rather than against final-task accuracy.

\paragraph{Selective prediction.} Abstention with a reject option dates to \citet{chow1970optimum}; risk--coverage analysis and the area under the risk--coverage curve formalize the accuracy--coverage trade-off for classifiers \citep{elyaniv2010foundations,geifman2017selective}. \citet{kamath2020selective} show that a model's own confidence is a poor abstention signal under domain shift for question answering, motivating a learned calibrator instead. LLM-specific confidence estimation includes self-evaluation prompts \citep{kadavath2022know} and semantic entropy over sampled generations \citep{kuhn2023semantic}. All of this work abstains the \emph{answering} model. We instead apply selective prediction to the \emph{evaluator}: the production assistant always answers, but \system\ may decline to score that answer when no sufficiently similar golden reference exists, which is a different decision (is this evaluation trustworthy?) from the one those methods address (is this answer trustworthy?) and admits different signals (retrieval margin $\Delta$, intent-conditional calibration) than answer-side confidence.

\end{document}